\documentclass{article}

\usepackage{PRIMEarxiv}

\usepackage[utf8]{inputenc} 
\usepackage[T1]{fontenc}    
\usepackage{hyperref}       
\usepackage{url}            
\usepackage{booktabs}       
\usepackage{amsfonts}       
\usepackage{nicefrac}       
\usepackage{microtype}      
\usepackage{lipsum}
\usepackage{amsmath}
\usepackage{graphicx}
\graphicspath{{media/}}     
\usepackage{subcaption}   
\usepackage{graphicx}     
\usepackage[inline]{enumitem} 
\usepackage{algorithm}
\usepackage{algpseudocode}  
\usepackage{booktabs}
\usepackage{rotating} 
\usepackage{multirow}
\usepackage{makecell}
\usepackage{threeparttable}
\title{OSC: Hardware Efficient W4A4 Quantization via Outlier Separation in Channel Dimension}

\author{
  Zhiyuan Zhang, Yanzhao Li, Zhiqiang Zou, Bai Du, Yupeng Sun, Hui Dong, Hui Wang \\
  Huawei Technology \\
    \texttt{\{zhangzhiyuan45, liyanzhao\}@huawei.com} \\
}

\begin{document}
\maketitle

\begin{abstract}
While 4-bit quantization is essential for high-throughput deployment of Large Language Models, activation outliers often lead to significant accuracy degradation due to the restricted dynamic range of low-bit formats. In this paper, we systematically investigate the spatial distribution of outliers and demonstrate a \textbf{token-persistent structural clustering effect}, where high-magnitude outliers consistently occupy fixed channels across  tokens. Building on this insight, we propose \textbf{OSC}, a hardware-efficient framework for outlier suppression. During inference, OSC executes a \textbf{dual-path computation} consisting of a low-precision 4bit General Matrix Multiplication (GEMM) path and a high-precision 16bit branch GEMM path. Specifically, OSC uses an offline \textbf{group-wise strategy} to identify the channels where outliers are located and then performs \textbf{structured sub-tensor extraction} to coalesce these scattered activation channels into a compact dense tensor online. This mechanism implements outlier protection through regularized and high-throughput GEMM operations, achieving a seamless fit with modern 4-bit micro-scaling hardware. Furthermore, for the inputs of $W_2$ where outlier clustering is less pronounced, we integrate an \textbf{fallback} strategy to FP8. Evaluation on Qwen3-8B and Qwen3-30B  restricts the average accuracy drop to \textbf{2.19} and \textbf{1.12} points, respectively. Notably, OSC is highly hardware-friendly, achieving a peak speedup of \textbf{1.78$\times$} over the W8A8 GEMM baseline on a modern AI accelerator.
\end{abstract}

\section{Introduction}

The unprecedented scaling of Large Language Models (LLM) has led to architectures with hundreds of billions of parameters \cite{deepseek_v3.2, llama3_1_report}, resulting in an explosion of computational demand and inference costs \cite{ai_report}. Given the prohibitive retraining overhead of Quantization-Aware Training (QAT) for large-scale models, Post-training Quantization (PTQ) remains the most widely adopted technology to alleviate deployment constraints. While 8-bit quantization is now widely adopted, pushing compression to 4-bit formats (e.g., W4A4) often triggers catastrophic accuracy degradation. This failure is primarily attributed to activation outliers---sparse but disproportionately high-magnitude values \cite{case_for_4bit}. Although modern group-wise micro-scaling formats like MXFP4 \cite{mxfp4}, NVFP4 \cite{nvfp4}, and HIF4 \cite{hif4} help mitigate general quantization noise, preserving model integrity against extreme outliers remains a formidable challenge.

Existing outlier suppression strategies typically rely on data-dependent dynamic sensing (e.g., LLM.int8() \cite{llmint8}, Top-P/K selection \cite{SpQR, Olive}) or global re-parameterization (e.g., SmoothQuant \cite{smoothquant}, QuaRot \cite{quarot}). While effective for accuracy recovery, these methods pay limited attention to the intrinsic requirements of fine-grained group-wise quantization and underlying hardware execution efficiency. Consequently, these methods introduce significant computational overhead, cause memory fragmentation, and complicate deterministic execution due to their runtime-dependent nature.

Prior work shows that activation outliers in LLMs cluster in a small subset of influential channels \cite{llmint8, smoothquant}. Based on these insights, this paper conduct a systematic, fine-grained statistical analysis of activation outliers across multiple modules. Our study reveals a \textbf{token-persistent structural clustering effect}: for the majority of transformer modules, such as Attention and FFN $W_1$/$W_3$ inputs, outliers consistently manifest in fixed channels within each quantization group, with clustering density often exceeding 60\%. Conversely, we identify specific "low-clustering" regions, most notably at the $W_2$ (down-projection) inputs, where outliers exhibit a more diffused distribution that defies static indexing.

Based on these insights, this paper propose \textbf{OSC}, a hardware-aligned \textbf{Offline Lookup Table} protection scheme. OSC leverages the structural priors of group-wise quantization (e.g., 32 elements for MXFP4) to perform static outlier isolation. By coalescing pre-indexed channels into a sub-tensor, OSC transforms scattered outlier protection into a regular, dense General Matrix Multiplication (GEMM) operation. For $W_2$ inputs that lack clustering stability, we employ a \textbf{hybrid-precision strategy}, selectively reverting these specific projections to the FP8 to ensure robust accuracy.

The main contributions of this paper are summarized as follows:
\begin{itemize}
    \item \textbf{Systematic Characterization of Outlier Heterogeneity:} Beyond identifying general channel outliers, this paper provide a fine-grained quantification of their spatial heterogeneity across computational blocks. By distinguishing between token-persistent structural clustering and stochastic diffusion, this paper establish a rigorous empirical foundation for position-specific, hardware-aligned quantization.
    
    \item \textbf{The OSC Framework:} This paper introduce OSC, a hardware-intrinsic outlier suppression scheme. By leveraging an offline lookup table and structured sub-tensor extraction, OSC performs outlier protection through regular, high-throughput GEMM operations. This approach effectively eliminates the runtime overhead and execution divergence typical of dynamic sensing methods.

    \item \textbf{Hybrid-precision Policy:} As a core component of OSC, OSC implement an adaptive precision policy that maps specific module profiles to optimal numerical formats. For $W_2$ inputs where static clustering is less pronounced, OSC utilize a selective \textbf{FP8 fallback}. This strategy balances aggressive 4-bit compression with high-fidelity protection, ensuring stability in sensitive MoE-based architectures.

    \item \textbf{Near-Lossless Accuracy and Hardware Efficiency:} This paper evaluate OSC on \textbf{Qwen3-8B} and \textbf{Qwen3-30B-A3B}, achieving near-lossless 4-bit inference accuracy across multiple benchmarks. Furthermore, this paper evaluate the practical efficiency of OSC through performance benchmarking of optimized hardware kernels, demonstrating significant speedups over baseline implementations on the high-performance accelerator architecture.
\end{itemize}

\section{Characterization of Outlier Clustering}
In this section, we provide a quantitative investigation into the spatial distribution of activation outliers to verify the proposed \textbf{token-persistent structural clustering effect}. Our analysis serves as the empirical foundation for the subsequent OSC protection scheme.

\subsection{Outliers Identification and Visual Evidence}
To enable outlier identification, we pre-define a layer-wise threshold $T_l$ during the offline profiling phase. We adopt a \textbf{layer-wise} rather than a global threshold because our empirical data suggest that activation magnitudes tend to escalate significantly as the model depth increases, particularly in the later stages of deep Transformer architectures. For an activation tensor $X_l \in \mathbb{R}^{S \times H}$ (where $S$ is the total number of tokens and $H$ is the hidden dimension, ignoring the batch dimension for clarity), the threshold is computed as:

\begin{equation}
T_l = \alpha \cdot \mathbb{E}[|X_l|] = \frac{\alpha}{S \cdot H} \sum_{i=1}^{S} \sum_{j=1}^{H} |x_{i,j}|
\end{equation}
where $\alpha$ is a scaling factor set to 5 in our study. This setting is determined empirically, ensuring that during grouping, most groups contain outliers without excessive inclusion.

\begin{figure*}[t]
  \centering
  \begin{minipage}[b]{0.32\textwidth}
    \centering
    \includegraphics[width=\textwidth, height=9.5cm, keepaspectratio=false]{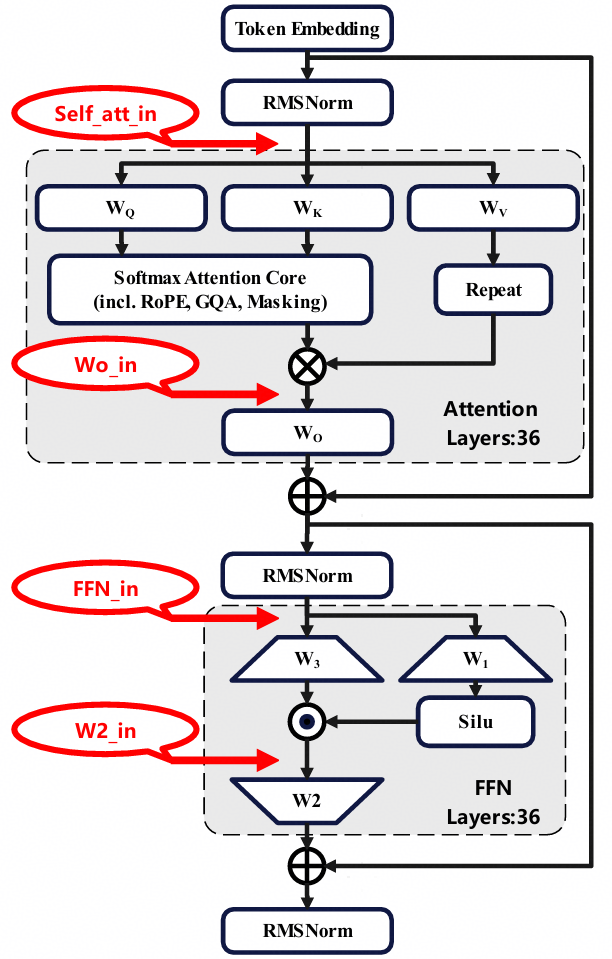}
    \caption{Data sampling visualization and the four monitored positions in Qwen3-8B.}
    \label{fig:fig1_Qwen3-8B}
  \end{minipage}
  \hfill
    \begin{minipage}[b]{0.65\textwidth}
      \centering
      \refstepcounter{figure}
      \label{fig:fig2_combined}
    
      \begin{subfigure}[b]{\textwidth}
        \centering
        \includegraphics[width=\textwidth, height=4cm]{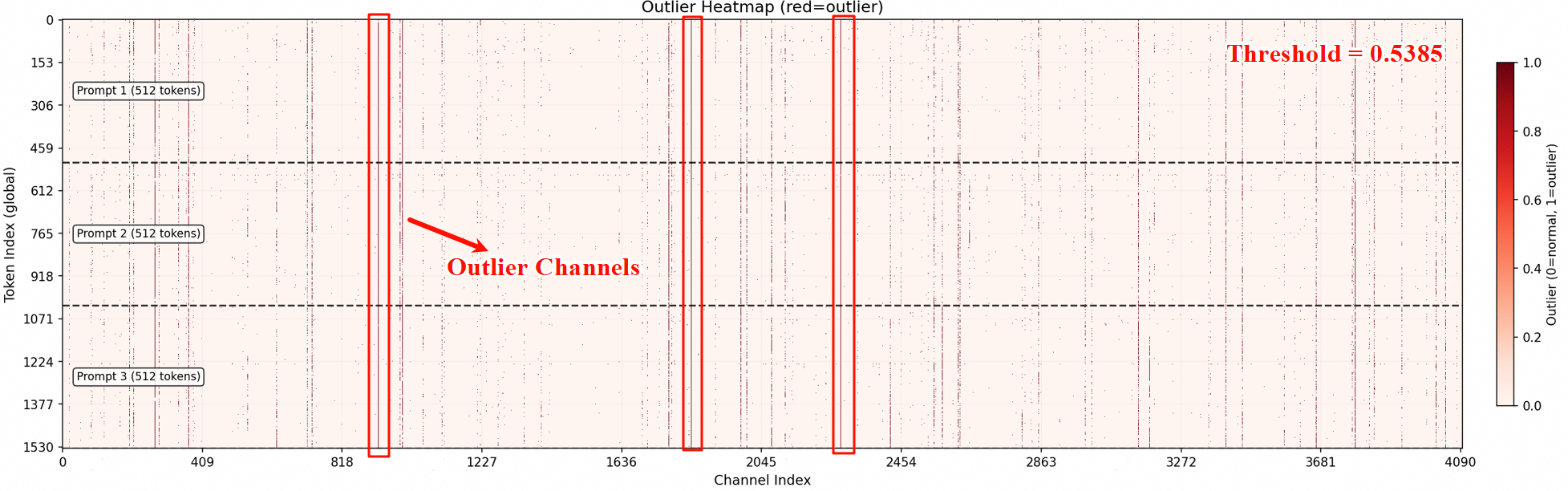}
        \caption{Outlier Boolean Mask ($|x| > T_l$)}
        \label{fig:outlier_mask}
      \end{subfigure}
    
      \vspace{0.4cm}
    
      \begin{subfigure}[b]{\textwidth}
        \centering
        \includegraphics[width=\textwidth, height=4cm]{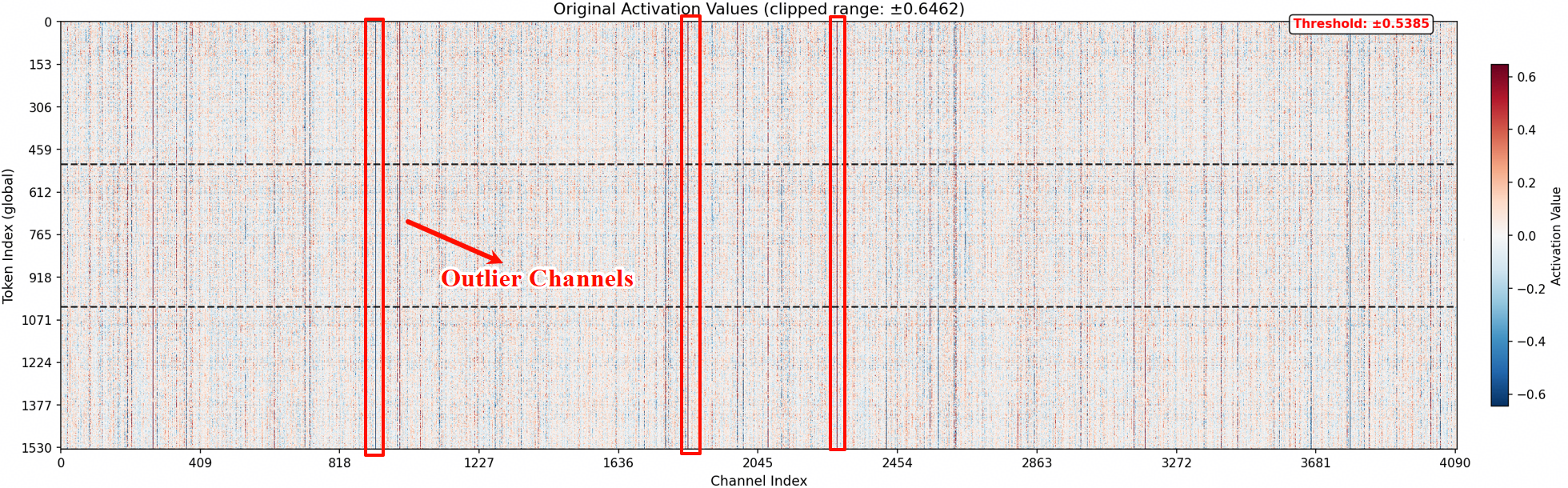}
        \caption{Raw Activation Magnitudes}
        \label{fig:magnitude_heatmap}
      \end{subfigure}
    
    \caption*{\textbf{Figure~\ref{fig:fig2_combined}:} Visualizing structural clustering in Qwen3-8B. The boolean mask (a) identifies tokens exceeding $T_l$, while the heatmap (b) displays the raw activation magnitudes, revealing that specific channels consistently harbor large-scale values across all tokens.}
    \end{minipage}
\end{figure*}

To investigate the distribution of activation outliers, we conduct a systematic analysis on the \textbf{Qwen3-8B} model. We specifically focus on four critical computational interfaces: 
    \begin{enumerate*}[label=(\arabic*)]
        \item inputs to the self-attention module ($W_Q$/$W_K$/$W_V$), 
        \item inputs to the $W_o$ projection, 
        \item inputs to the FFN ($W_1$/$W_3$), and 
        \item inputs to the $W_2$ projection, all of which are involved in GEMM operations during inference.
    \end{enumerate*}
The specific sampling points for Qwen3-8B \cite{qwen3_report} are illustrated in \textbf{Figure~\ref{fig:fig1_Qwen3-8B}}, which highlights the activation interfaces monitored during our analysis.

The spatial characteristics of the outliers are visualized in \textbf{Figure~\ref{fig:fig2_combined}}. Taking the $15^{th}$ self-attention input of Qwen3-8B as a representative case, we illustrate activation patterns aggregated over 3 distinct prompts (512 tokens each).

\begin{itemize}
    \item \textbf{Figure~\ref{fig:outlier_mask} (top):} Displays a binary mask $M \in \{0, 1\}^{S \times H}$, where $m_{i,j}=1$ (deep red) denotes an outlier ($|x_{i,j}| > T_l$) and $m_{i,j}=0$ (light red) denotes a normal value.
    \item \textbf{Figure~\ref{fig:magnitude_heatmap} (bottom):} Provides the corresponding magnitude heatmap of the activations across tokens and channels.
\end{itemize}

As illustrated in Figure~\ref{fig:fig2_combined}, the outliers exhibit a distinct \textbf{vertical stripe-like pattern}. This observation confirms that outliers are not stochastically distributed across the sequence but are persistently concentrated in specific channels, regardless of the input tokens, providing a reliable structural prior for offline lookup table construction.

\subsection{Quantifying Clustering Density}
To ensure compatibility with the group-wise microscaling formats optimized for modern AI accelerators, this paper partition the $H$ input channels into $K$ disjoint groups of size $G \in \{16, 32, 64\}$. Within this structural framework, we introduce the Clustering Density ($\mathcal{C}$) to quantify the spatial concentration of dominant outliers. By focusing on the token-wise representative outlier within each group, $\mathcal{C}$ serves as a direct proxy for the efficacy of static single-channel protection. The formal derivation of this metric follows a three-step process:

\begin{enumerate}
    \item \textbf{Token-wise Max Filtering:} For each token $s$ and each group $k$, we first identify the channel index $j^*_{s,k}$ that holds the maximum absolute magnitude within the group:
    \begin{equation}
    j^*_{s,k} = \arg\max_{j \in \{1, \dots, G\}} |x_{s,k,j}|
    \end{equation}
    
    \item \textbf{Outlier Sample Identification ($N_{total}^{(k)}$):} We define the set of "outlier-afflicted tokens" $\mathcal{S}_k$ for group $k$. A token $s$ is included in this set only if its maximum intra-group magnitude exceeds the threshold $T_l$:
    \begin{equation}
    \mathcal{S}_k = \{ s \mid |x_{s,k,j^*_{s,k}}| > T_l, s \in \{1, \dots, S\} \}
    \end{equation}
    The total count of such tokens is $N_{total}^{(k)} = |\mathcal{S}_k|$, representing the number of tokens that require outlier protection in this specific group.
    
    \item \textbf{Dominant Index Frequency ($N_{hit}^{(k)}$):} Among the tokens in $\mathcal{S}_k$, we identify the most frequent channel index $J^*_k$ across the sequence. The count of tokens whose primary outlier is captured by this specific index is:
    \begin{equation}
    N_{hit}^{(k)} = \max_{j \in \{1, \dots, G\}} \text{count}(\{ s \in \mathcal{S}_k \mid j^*_{s,k} = j \})
    \end{equation}
\end{enumerate}

The clustering density for group $k$ is thus defined as $\mathcal{C}_k = N_{hit}^{(k)} / N_{total}^{(k)}$. This metric provides a measure of the probability that protecting a single static channel will resolve the primary outlier threat for any given token. Intuitively, the final density for a computational block is the average across all $K$ groups: $\bar{\mathcal{C}} = \frac{1}{K} \sum_{k=1}^{K} \mathcal{C}_k$.

\textbf{Rationale for Metric Choice:} 
While the average clustering density $\bar{\mathcal{C}}$ may be numerically moderated by certain groups whose outliers are less extreme and thus exhibit less pronounced clustering, our extensive numerical analysis confirms that this metric remains a robust indicator of protection efficacy. We found that selecting only the most frequent outlier channel per group as the protection target yields near-optimal accuracy recovery while maintaining a highly regular hardware execution pattern. This structured approach is inherently robust: even if the offline profiling occasionally identifies a relatively low-magnitude value for protection (a false positive), the overall precision remains unaffected, as these values are simply processed in higher precision. Conversely, the high clustering density observed in critical groups ensures that catastrophic outliers (true positives), which are the primary drivers of quantization error, are rarely omitted from the lookup table. 

\begin{figure*}[t]
  \centering
  \begin{minipage}[b]{1\textwidth}
    \centering
    \includegraphics[width=\textwidth, height=5cm, keepaspectratio=false]{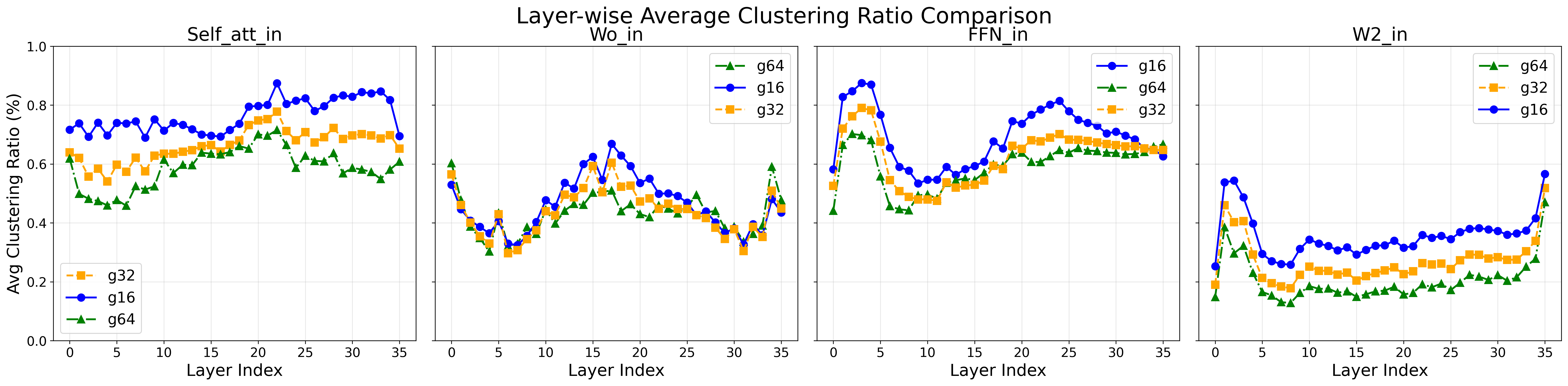}
    \caption{Distribution of outlier clustering density $\bar{\mathcal{C}}$ across \textbf{different layers} in Qwen3-8B. Each subplot represents a specific \textbf{module input position} (Attention, $W_o$, $W_1/W_3$, $W_2$), with individual lines within each plot depicting the clustering trends of \textbf{different groups of size ($G$)}.}
    \label{fig:fig3_qwen3_dist}
  \end{minipage}
\end{figure*}

\begin{table}[t]
\centering
\caption{Statistical summary of outlier clustering density in Qwen3-8B across different functional modules.}
\label{tab:qwen3_stats}
\begin{tabular}{lc}
\toprule
\textbf{Module Input} & \textbf{Clustering Density $\bar{\mathcal{C}}$ (\%)} \\
\midrule
Attention (self-attention inputs) & 60--80 \\
$W_o$ (output projection)         & 40--50 \\
$W_1$ / $W_3$ (FFN up-projection) & 60--70 \\
$W_2$ (down-projection)           & 20--35 \\
\bottomrule
\end{tabular}
\end{table}

\subsection{Spatial Heterogeneity and Generalizability}
The core motivation for our protection strategy lies in the spatial heterogeneity of outlier distributions, referring to the significant variance in clustering density across different functional modules and layer depths. Based on the metric $\bar{\mathcal{C}}$ defined in the previous section, we conducted a fine-grained statistical sweep of \textbf{Qwen3-8B} \cite{qwen3_report}. The empirical results, summarized in Table~\ref{tab:qwen3_stats} and Figure~\ref{fig:fig3_qwen3_dist}, reveal several critical patterns:

\begin{itemize}
    \item \textbf{High-Clustering Regions:} Inputs to the Attention ($W_Q$/$W_K$/$W_V$) and FFN-up ($W_1$/$W_3$) layers exhibit the highest density, typically ranging from \textbf{60\% to 80\%}. It is worth noting that this density interval is inherently sensitive to the quantization granularity, where a smaller group size $G$ (e.g., 16 vs. 64) increases the likelihood of capturing intra-group peaks due to the reduced search space. Their highly structured nature makes these regions ideal candidates for the static outlier suppression strategy of our OSC framework.
    \item \textbf{Moderate-Clustering Regions:} The $W_o$ (output projection) inputs show moderate clustering (\textbf{40\%--50\%}).
    \item \textbf{Low-Clustering Regions:} The $W_2$ (down-projection) inputs exhibit the lowest density (\textbf{20\%--35\%}), where outliers are diffused and less amenable to static indexing.
    \item \textbf{Layer-wise Stability:} Although $\bar{\mathcal{C}}$ may exhibit localized shifts—such as a sharp increase in specific early layers (e.g., $W_1$ in layers 1-4) reflecting architecture-specific characteristics—the clustering density for a given projection type remains within a stable, predictable interval across the majority of the model.
\end{itemize}

\textbf{Implications for OSC Design:} These empirical observations directly inform the tiered protection strategy of our framework:
\begin{itemize}
    \item \textbf{High-Clustering Regions ($\bar{\mathcal{C}} > 60\%$):} Modules such as Attention and $W_1/W_3$ are ideally suited for \textbf{static outlier indexing}. In these layers, protecting a single pre-identified channel per group captures the vast majority of outlier-afflicted tokens, yielding near-lossless accuracy recovery with minimal metadata.
    
    \item \textbf{Moderate-Clustering Regions ($35\% < \bar{\mathcal{C}} < 60\%$):} Although the clustering density in $W_o$ (output projection) is numerically lower than in FFN-up layers, our experiments demonstrate that \textbf{OSC's static suppression strategy} still provides a non-negligible accuracy gain. We attribute this to the fact that even at moderate density levels, the static index consistently captures the most extreme outliers that reside in the tails of the distribution. Thus, incorporating $W_o$ into the OSC's static suppression strategy path remains a cost-effective strategy for overall model stability.
    
    \item \textbf{Low-Clustering Regions ($\bar{\mathcal{C}} < 35\%$):} For "outlier-diffused" stages, particularly $W_2$ inputs, stochastic outliers defy static indexing. To maintain model integrity, the OSC framework \textbf{selectively configures these layers to the FP8 fallback path}. This strategy ensures robust accuracy by bypassing the limitations of static suppression in diffused regions, while preserving the 4-bit high-throughput advantage for the majority of the model.
\end{itemize}

\textbf{Cross-Model Consistency:} 
While we primarily focus on the statistical profile of Qwen3-8B in this section, we have independently verified these patterns on \textbf{Qwen3-30B-A3B} and \textbf{Llama3-8B} \cite{qwen3_report, llama3_1_report}. Although minor numerical variances exist in specific positons, the fundamental conclusion regarding structural clustering and spatial heterogeneity is consistent across these architectures, confirming the broad generalizability of our observations.

\section{The OSC Framework}
\label{sec:methodology}

The OSC framework operates on the dual principles of \textbf{static channel protection} and \textbf{adaptive fallback}. Based on the empirical observations in Section 2, we propose a three-stage approach to implement this strategy:
(1) \textbf{Offline Profiling} to identify high-density outlier channels and establish layer-wise thresholds; 
(2) \textbf{Static Outlier Suppression}, which integrates a dual-path execution pipeline into the quantized matrix multiplication, enabling hardware-efficient inference by decoupling outlier compensation from the low-bit computation path; and 
(3) \textbf{Adaptive Fallback}, which selectively configures layers with diffused outlier distributions to a higher-precision format (e.g., MXFP8). 
By dynamically assigning computational blocks to either the static suppression or the fallback path, OSC ensures optimal model integrity while preserving the high-throughput advantage of 4-bit operations.

\begin{algorithm}
\caption{OSC Profile-driven Indexing (Dimension-Aware)}
\label{alg:OSC_indexing}
\begin{algorithmic}[1]
\Require Calibration set $\mathcal{D}$, Positions $\mathcal{P}$, Layers $\mathcal{L}_{idx}$, Group size $G$, Threshold $T_l$
\Ensure Suppression Table $\mathcal{L} \in \mathbb{Z}^{P \times L \times K_{max}}$ \Comment{$K_{max}$ is the maximum groups across all $p$}
\State Initialize all $\mathcal{L}[p][l][k] \gets -1$ 
\For{each position $p \in \mathcal{P}$}
    \State $H_p \gets$ hidden\_dimension\_at($p$)
    \State $K_p \gets H_p / G$
    \For{each layer $l \in \{1, \dots, L\}$}
        \State $X \gets$ collect\_activations($\mathcal{D}, p, l$) \Comment{Shape: $S \times H_p$}
        \For{each group $k \in \{0, \dots, K_p-1\}$}
            \State $\mathcal{S}_{p,l,k} \gets \emptyset$ 
            \For{each token $s \in \{1, \dots, S\}$}
                \State $offset \gets k \times G$
                \State $j^* \gets \arg\max_{0 \le j < G} |x_{s, offset + j}|$
                \If{$|x_{s, offset + j^*}| > T_l$}
                    \State $\mathcal{S}_{p,l,k} \gets \mathcal{S}_{p,l,k} \cup \{j^*\}$
                \EndIf
            \EndFor
            \If{$\mathcal{S}_{p,l,k} \neq \emptyset$}
                \State $\mathcal{L}[p][l][k] \gets \text{mode}(\mathcal{S}_{p,l,k})$
            \EndIf
        \EndFor
    \EndFor
\EndFor
\end{algorithmic}
\end{algorithm}

\subsection{Static Suppression Table Construction}

To minimize runtime computational overhead, OSC identifies a pre-determined set of "outlier-prone" channels during a comprehensive offline calibration phase using representative data. The architectural core of OSC is a three-dimensional suppression lookup table $L \in \mathbb{Z}^{P \times L \times K}$, designed to store the structural priors of the model. Each entry $L_{p,l,k}$ stores a precise local index $j \in [0, G - 1]$, which identifies the single most statistically significant outlier channel within the $k$-th quantization group for a given layer and projection.

The indexing process operates in a hierarchical manner, as detailed in Algorithm~\ref{alg:OSC_indexing}. For each computational block, we first adaptively determine the group count $K_p$ based on the local hidden dimension. The core of the algorithm lies in the construction of the candidate set $\mathcal{S}_{p,l,k}$, which aggregates the indices of prominent outliers across all calibration tokens. By applying the mode operator to this set (see Line 17 of Algorithm~\ref{alg:OSC_indexing}), we extract the statistically dominant outlier channel, effectively transforming dynamic, token-dependent variance into a robust structural prior. If a group lacks consistently high-magnitude activations (i.e., the candidate set $\mathcal{S}$ is empty), it remains flagged with $-1$, signaling the hardware to bypass the suppression logic for that specific region.

\subsection{The OSC-Enhanced Quantization Procedure}
\label{sec:quant_proc}

Existing hardware-friendly quantization formats utilize micro-scaling by partitioning tensors into small blocks (e.g., $G=16, 32, 64$) to mitigate the impact of outliers. OSC augments this granularity by introducing a static extraction-and-restoration mechanism. By temporarily removing the most significant outlier in each group before calculating the scaling factors, OSC drastically reduces the dynamic range in quantization, thereby improving the quantization resolution for the remaining values.

As illustrated in Figure~\ref{fig:fig4_quant_workflow}, the OSC-enhanced quantization procedure is architected for seamless integration and high compatibility with standard micro-scaling frameworks. Most stages within this workflow---such as dimension alignment, tensor reshaping, and quantization. The primary innovation of OSC lies in the strategic insertion of two lightweight, hardware-friendly operations: \textbf{index-based lookup} from the offline table and \textbf{targeted element-wise zeroing} of identified outlier channels. 

For an input activation tensor $X \in \mathbb{R}^{B \times S \times H}$ residing at position $p$ within transformer layer $l$, the OSC-integrated quantization workflow is formally defined through the following hierarchical stages:
\begin{enumerate}
    \item \textbf{Dimension Alignment \& Reshaping (Standard)}: Identify the hidden dimension $H$ and reshape $X$ into a segmented $[ \dots, K, G ]$ format to match the hardware block size $G$ (e.g., 16, 32, or 64)\cite{zeroquant}.
    \item \textbf{OSC Index Retrieval (New)}: Fetch the pre-computed suppression vector $\mathcal{I} = \mathcal{L}[p][l] \in \mathbb{Z}^K$ from the offline lookup table.
    \item \textbf{OSC Outlier Zero-out (New)}: For each group $k$, if $\mathcal{I}[k] \ge 0$, extract the outlier at local index $j^* = \mathcal{I}[k]$ to a high-precision buffer $\mathcal{B}$ and set the corresponding entry in $X$ to zero. 
    \item \textbf{Quantization (Standard)}:  Calculate scaling factors and perform low-bit quantization on the modified
    tensor. Crucially, since the dominant outliers are removed in the previous step, the resulting scaling factors are
    much tighter, significantly reducing the overall quantization noise.
    \item \textbf{Outlier Restoration (Standard/Fused)}: Integrate the high-precision values from $\mathcal{B}$ back into the final computation path.
\end{enumerate}

By only introducing an $O(K)$ lookup and assignment overhead—where the group count $K$ is typically two orders of magnitude smaller than the total hidden dimension $H$—OSC achieves a superior accuracy-efficiency trade-off compared to existing dynamic outlier detection schemes. Unlike dynamic methods that require real-time sorting or thresholding, OSC's static nature allows it to bypass expensive runtime searching entirely.

The comprehensive algorithmic logic is detailed in Algorithm~\ref{alg:OSC_quant}. This procedure maintains a remarkably low computational complexity of $O(K)$, introducing negligible latency overhead compared to the subsequent compute-intensive GEMM operations. Consequently, the total execution time remains dominated by the high-throughput matrix multiplication, allowing OSC to provide robust outlier protection with near-zero performance degradation.

\begin{algorithm}[H]
\caption{OSC-Enhanced Quantization Workflow}
\label{alg:OSC_quant}
\begin{algorithmic}[1]
\Require Input tensor $X \in \mathbb{R}^{S \times H}$, Block size $G$, Suppression indices $\mathcal{I} = \mathcal{L}[p][l]$
\Ensure Quantized base tensor $Q_{base}$, Outlier buffer $\mathcal{B}$
\State $K \gets H / G$
\State $X_{reshaped} \gets \text{reshape}(X, [S, K, G])$
\State $\mathcal{B} \gets \text{zeros}(S, K)$ \Comment{Temporary high-precision buffer}
\For{each group $k \in \{0, \dots, K-1\}$}
    \State $j^* \gets \mathcal{I}[k]$
    \If{$j^* \ge 0$} \Comment{Check for suppression flag}
        \State $\mathcal{B}[:, k] \gets X_{reshaped}[:, k, j^*]$ \Comment{Backup original high-precision values}
        \State $X_{reshaped}[:, k, j^*] \gets 0$ \Comment{Zero-out to reduce group dynamic range}
    \EndIf
\EndFor
\State $Scale \gets \text{CalculateScale}(X_{reshaped}, G)$ \Comment{Standard micro-scaling scale calculation}
\State $Q_{base} \gets \text{Quantize}(X_{reshaped}, Scale)$ 
\State \Return $\{Q_{base}, \mathcal{B}, \mathcal{I}\}$
\end{algorithmic}
\end{algorithm}

\subsection{Hybrid-Precision Matrix Computation and Hardware Execution}
\label{sec:hybrid_comp}

In the OSC-enhanced quantization framework, the inference engine performs a dual-path hybrid-precision GEMM. This section outlines the computational fusion logic and hardware-level Strategy mechanisms that enable high-throughput execution, balancing low-bit and high-precision operations efficiently.

\subsubsection{Hybrid Computational Formulation}
The final output $Y$ is reconstructed by merging the low-bit quantized result with the high-precision outlier compensation \cite{llmint8}, as illustrated in Figures~\ref{fig:fig4_quant_workflow}. For a given input activation $X$ and weight matrix $W$, the operation can be expressed as:

\begin{figure*}[t]
  \centering
  \begin{minipage}[b]{1\textwidth}
    \centering
    \includegraphics[width=\textwidth, height=6cm, keepaspectratio=false]{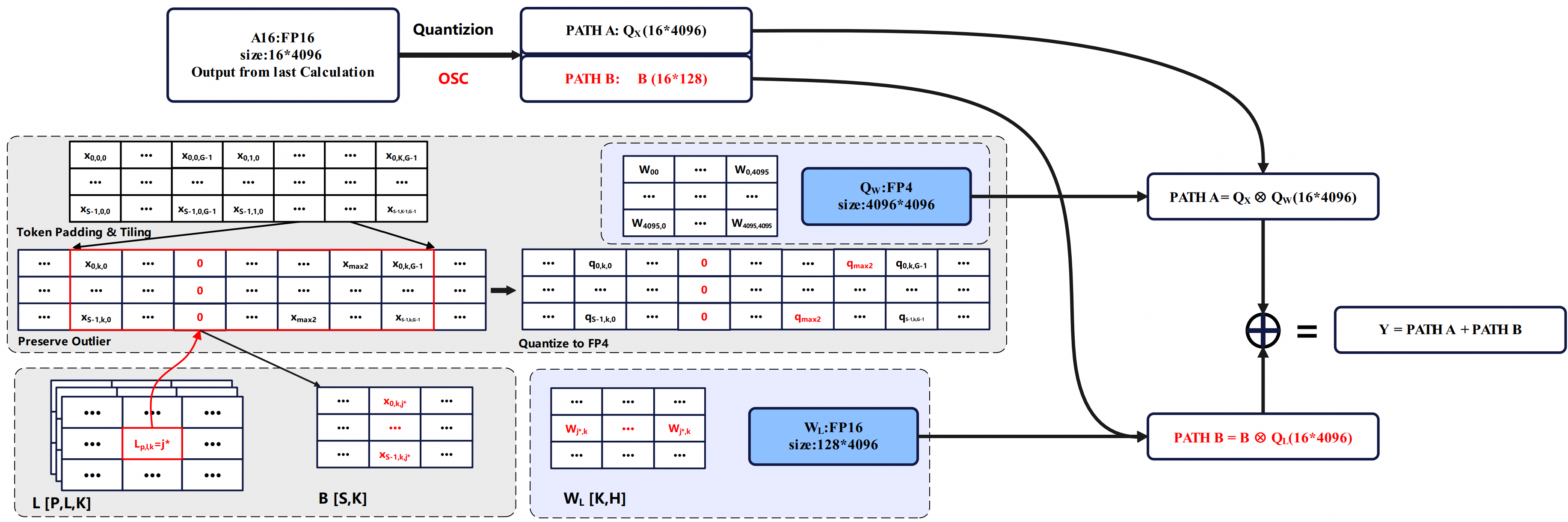}
    \caption{OSC-Enhanced Quantization Workflow ($G$=32).}
    \label{fig:fig4_quant_workflow}
  \end{minipage}
\end{figure*}

\begin{equation}
\label{eq:dual_path}
Y = \underbrace{(Q_{X} \otimes Q_{W}) \cdot \mathcal{S}}_{\text{Path A: Low-bit Base}} + \underbrace{\mathcal{B} \cdot \mathcal{W}_{\mathcal{L}}}_{\text{Path B: High-precision Bypass}}
\end{equation}

In this formulation, $Q_{X}$ and $Q_{W}$ represent the 4-bit quantized activation and weight tensors. The core efficiency of OSC resides in Path B, which is mathematically equivalent to the outlier partial product but formulated as a dense, hardware-friendly matrix multiplication:
\begin{itemize}
\item $\mathcal{B} \in \mathbb{R}^{S \times K}$ is a compact activation sub-tensor where each column $k$ is extracted from the original activation $X$ using the pre-computed index $\mathcal{L}[p][l][k]$.
\item $\mathcal{W}_{L} \in \mathbb{R}^{K \times H}$ is a structured weight sub-matrix composed of rows gathered from the high-precision weights $W_{fp}$ according to the same static lookup table $\mathcal{L}$.
\end{itemize}
By coalescing scattered outliers into these continuous memory blocks, OSC transforms irregular compensation into a standard GEMM, eliminating the need for runtime dequantization or dynamic branching.


\begin{table}[t]
\centering
\caption{Hardware Efficiency: Evaluated on a modern architecture with a 1:2:4 (FP16:FP8:FP4) performance ratio, OSC achieves a 1.5$\times$–1.88$\times$ speedup over the W8A8 baseline.}
\label{tab:efficiency}
\renewcommand{\arraystretch}{1.3} 
\begin{tabular}{c|cc}
\toprule
 {\textbf{Group Size ($G$)}} & \textbf{$M$ = 16-64} & \textbf{$M \ge$ 128}\\
\midrule
\textbf{16}  & 1.50$\times$-1.59$\times$ & 1.6$\times$ \\
\textbf{32}  & 1.64$\times$-1.76$\times$ & 1.78$\times$ \\
\textbf{64}   & 1.72$\times$-1.86$\times$ & 1.88$\times$ \\
\bottomrule
\end{tabular}
\end{table}

\subsubsection{Hardware Data Coordination and Efficiency Analysis}
The primary hardware advantage of OSC lies in its \textbf{structural determinism}, which enables highly optimized kernel execution on AI accelerators. To minimize latency caused by outlier suppression and maximize throughput, as shown in Figure~\ref{fig:fig4_quant_workflow}, OSC implements several hardware-level data movement and Coordination strategies:

\begin{itemize}
    \item \textbf{Zero-Overhead Offline Weight Quantization}: Weights are pre-quantized into FP4 formats (e.g., $W_{4bit}[4096, 4096]$).In Path B, token-wise column extraction is performed, where the corresponding weight rows are pre-stored in memory as high-precision sparse weight rows (e.g., $W_{16bit}[128, 4096]$). The suppression indices $\mathcal{L}$ are pre-determined offline, allowing the hardware to pre-calculate the memory offsets for these high-precision sparse weight rows during kernel loading. This eliminates the need for runtime searching or dynamic branching, significantly reducing overhead.
    \item \textbf{Dual-Path Activation Flow}: High-precision activations from the preceding layer (e.g., $A[16, 4096]$) are split into $A_{4bit}$ for Path A and a sparse $A_{16bit}$ buffer for Path B. In a 32-to-1 protection scheme of MXFP4 format, Path B only processes a fractional subset of the total data.
    \item \textbf{Adaptive Resource Dispatching}: Depending on the hardware's resource configuration, Path B can be dispatched to high-precision Matrix Units (Cube/Tensor cores).
\end{itemize}

These strategies together ensure that the OSC framework executes with minimal latency and high computational throughput. The efficiency gains of OSC are driven by several key structural and architectural optimizations:
\begin{enumerate}
    \item \textbf{Reduced Arithmetic and Memory Footprint}: The 4-bit base computation (Path A) halves memory traffic compared to 8-bit formats and leverages high-throughput 4-bit matrix units, directly cutting cycle counts.
    
    \item \textbf{Structured Outlier Extraction}: Unlike dynamic schemes that often suffer from suboptimal hardware utilization, OSC coalesces irregular sparse accesses into dense sub-tensor operations. This transformation enables efficient memory coalescing and eliminates control-flow divergence, allowing the hardware to maintain a high execution-to-stall ratio and fully saturate the throughput of matrix-multiplication units.    
    
    \item \textbf{Negligible Indexing Overhead}: The static indexing table $\mathcal{L}$ is exceptionally compact, as it only stores local group indices. With a memory footprint of only a few kilobytes, the table is effectively "transparent" to the memory hierarchy, residing entirely within the L1/L2 cache to avoid high-latency global memory fetches.
    
    \item \textbf{Minimal Arithmetic Branch Cost}: Path B introduces only one additional floating-point MAC (Multiply-Accumulate) operation for each group of size $G$. While the theoretical ratio of operations is $1/G$, the actual impact on cycles is modulated by the throughput ratios of the target architecture, assuming a peak performance ratio of 1:2:4 for FP16:FP8:FP4. Even accounting for this $4 \times$ throughput difference between the high-precision branch and the 4-bit base path, the arithmetic overhead remains minimal (e.g., $\approx 12.6\%$ for $G=32$), ensuring that the precision recovery does not bottleneck the primary 4-bit execution flow. 

\end{enumerate}
The performance metrics in Table~\ref{tab:efficiency} are evaluated across representative GEMM dimensions $\mathbf{A} \times \mathbf{B}$, where $\mathbf{A} \in \mathbb{R}^{M \times K}$ and $\mathbf{B} \in \mathbb{R}^{K \times N}$ denote the activation and weight matrices, respectively. By sweeping the classical combinations of $K$ and $N$ through $\{512, 768, 2048, 4096, 12288\}$—which align with the architectural configurations of Qwen3-8B and Qwen3-30B-A3B \cite{qwen3_report}—while fixing the batch size $M$, the results reveal two distinct hardware-bottleneck regimes:

\begin{itemize}
    \item \textbf{Memory-bound Regime ($M \in [16, 64]$)}: At smaller $M$ dimensions (typically corresponding to the \textit{decode} stage with small batch sizes), the arithmetic intensity is low, and performance is primarily limited by memory bandwidth. In this regime, OSC maintains a consistent \textbf{1.50$\times$--1.86$\times$} speedup over the W8A8 baseline. This confirms that by compressing weights to 4-bit, OSC effectively halves the weight-loading volume, directly alleviating the pressure on the memory interface.
    
    \item \textbf{Compute-bound Regime ($M \ge 128$)}: As the $M$ dimension increases (characteristic of the \textit{prefill} stage or large-batch inference), the data reuse ratio rises significantly, shifting the bottleneck entirely to the peak throughput of the CUBE computational units. Here, the speedup scales notably, reaching a peak of \textbf{1.88$\times$} at $G=64$. This proves that OSC's deterministic dual-path execution eliminates runtime-dependent stalls and control-flow divergence, allowing the hardware to fully saturate the 4-bit computational potential of high-throughput AI accelerators.
\end{itemize}

Compared to the W8A8 baseline, OSC achieves significant cycle savings (up to \textbf{43.8\%} reduction in cycles for $M\ge 128$ and $G=32$ ), primarily due to the transition to 4-bit arithmetic and the elimination of runtime dequantization. By following a fixed, deterministic execution path, OSC avoids the memory continuity issues and frequent data transpositions inherent in dynamic Top-K/Top-P solutions, providing a robust and high-throughput solution for low-bit LLM deployment.

\subsection{Hybrid-precision Adaptive Fallback Strategy}
\label{sec:adaptive_fallback}

While the OSC framework provides efficient and accurate solutions for most model modules, empirical analysis (Section 2.3) reveals that its effectiveness depends heavily on the spatial clustering of outliers. In some positions, particularly the $W_2$ (Down-projection) inputs in the FFN block, outliers show a more dispersed distribution than in other positions like self-attention or $W_1/W_3$. 

For these regions with low clustering of outliers, a single static index per group is insufficient. To accommodate these dispersed outliers without introducing excessive architectural complexity, we propose the \textbf{hybrid-precision Fallback} strategy:

\begin{itemize}
    \item \textbf{Standard method}: Use the 4-bit OSC framework for efficient compression and computational throughput.
    \item \textbf{$W_2$-specific Fallback}: For the $W_2$ inputs, switch the quantization format to \textbf{FP8} (or equivalent 8-bit micro-scaling formats). This provides the dynamic range required to handle dispersed outliers while maintaining a hardware-friendly micro-scaling structure.
\end{itemize}

The targeted fallback to FP8 for the $W_2$ inputs ensures that the system achieves a balance between aggressive bit-reduction and robust accuracy preservation across the entire model. As demonstrated in the experimental section, this adaptive strategy provides a significant improvement in accuracy without sacrificing hardware efficiency. Additionally, this strategy can be adjusted based on the specific characteristics of different model architectures. For example, the fallback can be fine-tuned to specific layers, allowing for a more granular adjustment that optimizes both computational efficiency and model accuracy.

\section{OSC: Experiment Results}
\label{sec:experiments}

\subsection{Experimental Setup}

\textbf{Models and Architectures:} We evaluate OSC on two representative Large Language Models from the Qwen3 family \cite{qwen3_report}:

\begin{itemize}
    \item \textbf{Dense Model}: Qwen3-8B (32 layers).
    \item \textbf{MoE Model}: Qwen3-30B-A3B (48 layers, 128 experts with 8 Activated Experts).
\end{itemize}

\textbf{Datasets and Tasks:} To assess the zero-shot and few-shot reasoning capabilities, we use a comprehensive suite of benchmarks: MMLU \cite{MMLU} (5-shot), GSM8K \cite{GSM8K} (8-shot), and ARC \cite{ARC} (Challenge/Easy, 25-shot).

\textbf{Implementation Details:} The efficacy of the OSC framework is assessed under the MXFP4 micro-scaling regime at a granularity of $G=32$. The static suppression lookup tables are generated using a calibration set of 3 sequences (512 tokens each) randomly sampled from the Pile dataset. For the $W_2$ inputs in FFN blocks, we implement a fallback strategy to MXFP8 \cite{mxfp4} to maintain numerical stability against diffused outlier distributions.

\begin{table}[!t]
\centering
\small
\setlength{\tabcolsep}{3.5pt}
\begin{threeparttable}
\caption{Accuracy Results of Qwen3 models under different quantization methods and strategies (scores with accuracy drop relative to W16A16 baseline; Mean column shows average score and average drop over four tasks)}
\label{tab:qwen3_combined}
\begin{tabular}{>{\centering\arraybackslash}p{1.2cm} l c c c c | c}
\toprule
\addlinespace[-\aboverulesep]
\midrule
Model & Method-Strategy & MMLU & GSM8K & ARC-C & ARC-E & Mean \\
\midrule
\multirow{6}{*}{\rotatebox[origin=c]{90}{Qwen3-8B}}
 & W16A16-Direct      & 74.97 (0.00) & 88.48 (0.00) & 67.58 (0.00) & 87.96 (0.00) & 79.75 (0.00) \\
 & W8A8-Direct        & 74.61 (-0.36) & 87.64 (-0.83) & 67.58 (0.00) & 87.75 (-0.21) & 79.40 (-0.35) \\
\cmidrule(lr){2-7}
 & MXFP4-Direct       & 68.24 (-6.73) & 80.44 (-8.04) & 61.43 (-6.14) & 84.51 (-3.45) & 73.66 (-6.09) \\
 & MXFP4-$W_2$FP8\tnote{1}        & 70.31 (-4.66) & 83.85 (-4.62) & 64.42 (-3.16) & 86.11 (-1.85) & 76.17 (-3.58) \\
 & MXFP4-$W_2$FP8+Dyn\tnote{2}     & 72.64 (-2.33) & 86.20 (-2.27) & 63.99 (-3.58) & 86.95 (-1.01) & 77.45 (-2.30) \\
 & MXFP4-Dynamic\tnote{3}      & 72.26 (-2.71) & 85.29 (-3.18) & 63.99 (-3.58) & 86.66 (-1.30) & 77.05 (-2.70) \\
 & \textbf{MXFP4-OSC}\tnote{4}  & \textbf{72.40 (-2.56)} & \textbf{85.82 (-2.65)} & \textbf{65.10 (-2.47)} & \textbf{86.91 (-1.05)} & \textbf{77.56 (-2.19)} \\
\cmidrule(lr){1-7}
\cmidrule(lr){1-7}
\multirow{6}{*}{\rotatebox[origin=c]{90}{Qwen3-30B}}
 & W16A16-Direct      & 82.03 (0.00) & 89.99 (0.00) & 72.78 (0.00) & 90.57 (0.00) & 83.84 (0.00) \\
 & W8A8-Direct        & 81.98 (-0.06) & 89.69 (-0.30) & 72.87 (+0.09) & 90.61 (+0.04) & 83.79 (-0.05) \\
\cmidrule(lr){2-7}
 & MXFP4-Direct       & 78.23 (-3.80) & 88.02 (-1.97) & 70.31 (-2.47) & 88.85 (-1.73) & 81.35 (-2.49) \\
 & MXFP4-$W_2$FP8\tnote{1}        & 78.93 (-3.10) & 88.17 (-1.82) & 71.50 (-1.28) & 89.35 (-1.22) & 81.99 (-1.85) \\
 & MXFP4-$W_2$FP8+Dyn\tnote{2}     & 79.91 (-2.12) & 88.78 (-1.21) & 73.21 (+0.43) & 90.19 (-0.38) & 83.02 (-0.82) \\
 & MXFP4-Dynamic\tnote{3}      & 79.78 (-2.25) & 89.16 (-0.83) & 71.76 (-1.02) & 90.57 (0.00) & 82.82 (-1.03) \\
 & \textbf{MXFP4-OSC}\tnote{4}  & \textbf{79.28 (-2.75)} & \textbf{89.01 (-0.99)} & \textbf{72.61 (-0.17)} & \textbf{89.98 (-0.59)} & \textbf{82.72 (-1.12)} \\
\addlinespace[-\aboverulesep]
\bottomrule
\end{tabular}
\begin{tablenotes}
\item[1] $W_2$FP8: Selective fallback where $W_2$ weights in FFN blocks use MXFP8. 
\item[2] $W_2$FP8+Dyn: Combines FP8 fallback on $W_2$ inputs with dynamic protection on all other positions.
\item[3] Dynamic: Per-group runtime max-value detection to guide mixed-precision.
\item[4] OSC: Combines $W_2$FP8 fallback with static outlier suppression on high-clustering positions.
\end{tablenotes}
\end{threeparttable}
\end{table}

\subsection{Main Results}
\label{subsec:main_results}

Table~\ref{tab:qwen3_combined} presents a comprehensive ablation study of the OSC framework against several baseline quantization and protection strategies. This evaluation aims to decouple the accuracy gains provided by the $W_2$ fallback from the core static outlier suppression mechanism.

\textbf{Effectiveness of Static Outlier Suppression.} The results demonstrate that while the $W_2$FP8 fallback (MXFP4-$W_2$FP8) recovers a significant portion of accuracy compared to direct quantization—reducing the average drop from 6.09 to 3.58 points on Qwen3-8B—it remains insufficient for near-lossless inference. By integrating our static suppression mechanism, OSC (MXFP4-OSC) further reduces the accuracy loss to 2.19 points. This confirms that OSC’s ability to identify and suppress high-clustering outlier channels across non-$W_2$ layers provides critical precision recovery that cannot be achieved by layer-wise fallback alone.

\textbf{OSC vs. Optimized Dynamic Protection.} To further validate the efficacy of OSC, we implemented an idealized Dynamic Protection baseline as a competitive control group. This strategy performs per-group maximum value detection at runtime to guide mixed-precision execution. Although this dynamic approach is theoretically flexible, OSC achieves superior accuracy (2.19 vs. 2.70 drop on 8B) while eliminating the substantial hardware overhead associated with runtime search. On the Qwen3-30B model, OSC achieves an average loss of 1.12 points, proving its robustness across different model scales.

\textbf{Hardware-Efficiency Trade-offs.} A key advantage of OSC over the Dynamic control group lies in its hardware-friendly execution. Unlike dynamic protection, which requires complex control flow and irregular memory access to handle outliers on the fly, OSC’s suppression decisions are determined entirely during the offline calibration phase. As shown in Table \ref{tab:efficiency}, this allows OSC to extract significant architectural benefits even under extremely constrained selection budgets, achieving peak speedups of 1.78$\times$.

Overall, these results demonstrate that OSC provides a robust, hardware-friendly solution for mitigating accuracy loss in 4-bit quantized LLMs, enabling near-lossless inference across diverse model families and task types.

\section{Related Work}
\label{sec:related_work}

\subsection{Systematic Characterization of Activation Outliers}
Prior research has established that activation outliers in LLMs exhibit significant clustering effects, predominantly residing in a small subset of persistent channels \cite{llmint8,smoothquant, massive_activations}. Early studies like LLM.int8() \cite{llmint8} and SmoothQuant \cite{smoothquant} identified that activation outliers in LLMs are typically confined to specific influential channels. While foundational, they predominantly treated outliers as monolithic, channel-wide anomalies. In contrast, this work conducts a fine-grained, quantitative statistical analysis of outlier spatial distribution and clustering density within quantization groups. We uncover a \textbf{token-persistent structural clustering effect} and highlight \textbf{spatial heterogeneity} across computational stages (e.g., Attention vs. FFN $W_2$ inputs), providing a rigorous statistical basis for deterministic hardware acceleration.

\subsection{Evolution of Outlier Mitigation Strategies}
The challenge of quantizing LLMs with extreme outliers has driven several mitigation strategies, broadly categorized as follows:

\textbf{Reparameterization and Orthogonal Transformations:} Methods like SmoothQuant \cite{smoothquant} and AWQ \cite{awq} migrate outlier magnitudes from activations to weights via scaling and shifting. However, this merely re-distributes rather than eliminates outliers, still stressing the narrow dynamic range of sub-8-bit formats. For instance, SmoothQuant can incur a significant accuracy degradation of up to 15 points on Llama2-70B in W4A4 scenarios \cite{spinquant}. Techniques such as QuaRot \cite{quarot} and FlatQuant \cite{flatquant} apply orthogonal transformations to "smear" outliers across feature dimensions, with QuaRot reporting an accuracy loss of approximately 3.5 points. Most recently, SpinQuant \cite{spinquant} advances this paradigm by employing learned rotations to further smooth the activation distribution, narrowing the gap to 1.9 points, while FlatQuant \cite{flatquant} demonstrates a loss of only 1.4 points. While theoretically elegant, such reparameterization methods introduce additional linear transformation overhead during inference and often disrupt hardware-friendly memory layouts optimized for modern AI accelerators.

\textbf{Dynamic Extraction and Sparse Handling:} Approaches like LLM.int8() \cite{llmint8}, SpQR \cite{SpQR}, and Atom \cite{atom} adopt runtime strategies—such as dynamic rescaling or search-and-extract—to isolate salient values. Although achieving near-lossless accuracy, their data-dependent execution leads to non-deterministic control flows and unstructured memory access. This prevents the underlying GEMM kernels from fully saturating the peak throughput of dense matrix multiplication units on high-performance accelerators.

\textbf{Static Retention and Structural Approaches:} Recognizing the overhead of dynamic search, a recent concurrent work, TetraJet-v2 \cite{tetrajetv2}, explores static outlier retention to mitigate activation oscillations during FP4 training. While both OSC and TetraJet-v2 leverage the structural stability of outliers via static indexing, their objectives differ fundamentally. TetraJet-v2 focuses on stabilizing training dynamics, whereas OSC is engineered for \textbf{inference-time throughput optimization}. Furthermore, OSC introduces a \textbf{deterministic dual-path mechanism} at the group level, \textit{which is tightly coupled with the micro-scaling granularity of the underlying data format}. This design explicitly addresses the \textbf{spatial heterogeneity} of outliers---such as the diffused nature of $W_2$ inputs---representing a critical nuance for maintaining accuracy in large-scale MoE models.

\subsection{Summary: Bridging Algorithms and Hardware Execution}
In summary, while prior works excel in hardware-agnostic accuracy recovery or general-purpose smoothing, OSC bridges the gap between fine-grained statistical insights and deterministic hardware execution. By transforming the stochastic problem of outlier handling into structured matrix multiplications via static indexing, OSC achieves high-precision recovery without sacrificing the compute-bound performance of modern AI accelerators.

\section{Conclusion}
This paper introduces a hardware-friendly static outlier suppression framework (\textbf{OSC}) that enables accurate and efficient 4-bit inference for Large Language Models. Our key insight is the discovery of spatial heterogeneity in activation outliers: while most layers exhibit highly clustered outlier channels amenable to static indexing, specific regions—particularly $W_2$ inputs—show dispersed distributions that require higher precision. Leveraging this insight, OSC adopts a hybrid-precision design: for layers with strong outlier clustering, it uses static indexing and dual-path execution to protect outliers with minimal overhead; for outlier-diluted regions such as $W_2$ inputs, it applies a selective FP8 fallback to ensure robustness. This strategy harmonizes outlier protection with the deterministic nature of GEMM operations, preserving model integrity while maximizing the throughput potential of modern AI accelerators.

Extensive experiments on the Qwen3-8B and Qwen3-30B models demonstrate that OSC consistently outperforms direct 4-bit quantization and provides superior accuracy recovery compared to specialized fallback and dynamic protection strategies. In our ablation study using the MXFP4 format, OSC reduces the average accuracy drop to approximately 2.19 points on the dense model and 1.12 points on the MoE architecture, effectively bridging the gap to the full-precision baseline. Performance evaluations on a modern AI accelerator further confirm the efficiency of OSC, which achieves speedups of  1.64$\times$--1.78$\times$ for MXFP4 over the W8A8 GEMM baseline. Similar precision and performance gains are also consistently observed across NVFP4 and HIF4 formats, highlighting the robustness of the OSC framework. These results position OSC as a practical and scalable solution for the deployment of 4-bit LLMs in production environments.

\textbf{Limitations:} Despite the promising results, this work has certain limitations. First, our evaluation primarily relies on comprehensive accuracy simulations and analytical performance modeling based on high-performance AI accelerators. While these models are grounded in peak throughput ratios (e.g., $1:2:4$ for FP16:FP8:FP4), the end-to-end latency in a physical deployment may be influenced by additional system-level factors. Furthermore, our adaptive fallback strategy is currently implemented at a relatively coarse-grained layer-type level. Specifically, while we identify $W_2$ inputs as outlier-diluted regions, OSC applies the FP8 fallback to all $W_2$ layers uniformly. This lacks a fine-grained selection mechanism that could potentially identify specific $W_2$ layers or even sub-groups that still possess enough structural clustering to benefit from static channel protection, thereby missing opportunities for further hardware acceleration.

\textbf{Future Work:} Future work will explore adaptive layer-wise fallback, extend the framework to additional model families and hardware platforms, and investigate its integration with quantization-aware training (QAT) to further enhance accuracy.

\bibliographystyle{unsrt}  
\bibliography{references}  

\end{document}